\def\year{2020}\relax
\documentclass[letterpaper]{article} 
\usepackage{aaai20}  
\usepackage{times}  
\usepackage{helvet} 
\usepackage{courier}  
\usepackage[hyphens]{url}  
\usepackage{graphicx} 
\usepackage{graphicx}  
\usepackage{latexsym}
\usepackage{booktabs} 
\usepackage{amsmath}
\usepackage{amsfonts}
\usepackage{bm}
\usepackage{array}
\usepackage{enumitem}
\usepackage{needspace}
\usepackage{adjustbox}
\usepackage{tikz}
\usepackage{pgfplots}
\usepackage{multirow}
\usepackage{autobreak}
\usepackage{makecell}
\usepackage{xcolor}
\usepackage{amssymb}
\usepackage{import}
\usepackage{subcaption}
\usepackage{nicefrac}       
\usepackage{microtype}      
\usepackage{romannum}
\usepackage{tabularx}
\usepackage{dcolumn}

\newcommand\setrow[1]{\gdef\rowmac{#1}#1\ignorespaces}
\newcommand\clearrow{\global\let\rowmac\relax}
\clearrow

\newcommand{\citet}[1]{\citeauthor{#1} \shortcite{#1}} 
\newcommand{\citep}{\cite}

\title{Learning from Easy to Complex: Adaptive Multi-curricula Learning for Neural Dialogue Generation}

\author{
Hengyi Cai,$^{\dagger,\S}$\thanks{Work done at Data Science Lab, JD.com.}
Hongshen Chen,$^\ddagger$
Cheng Zhang,$^\dagger$
Yonghao Song,$^\dagger$\\
{\Large\bf
Xiaofang Zhao,$^\dagger$
Yangxi Li,$^{\diamond}$
Dongsheng Duan,$^{\diamond}$
Dawei Yin$^\ddagger$} \\
$^\dagger${Institute of Computing Technology, Chinese Academy of Sciences} \\
$^\S${University of Chinese Academy of Sciences, Beijing, China} \\
$^\ddagger${Data Science Lab, JD.com} \\
$^{\diamond}${National Computer network Emergency Response technical Team/Coordination Center of China}\\
{caihengyi@ict.ac.cn, ac@chenhongshen.com, \{zhangcheng, songyonghao, zhaoxf\}@ict.ac.cn,}\\
{liyangxi@outlook.com, dds@cert.org.cn, yindawei@acm.org}
}

\begin{document}

\maketitle

\begin{abstract}
 Current state-of-the-art neural dialogue systems are mainly data-driven and are trained on human-generated responses.
 However, due to the subjectivity and open-ended nature of human conversations, the complexity of training dialogues varies greatly. 
 The noise and uneven complexity of query-response pairs impede the learning efficiency and effects of the neural dialogue generation models. 
 What is more, so far, there are no unified dialogue complexity measurements, and the dialogue complexity embodies multiple aspects of attributes---specificity, repetitiveness, relevance, etc.
 Inspired by human behaviors of learning to converse, where children learn from easy dialogues to complex ones and dynamically adjust their learning progress, in this paper, we first analyze five dialogue attributes to measure the dialogue complexity in multiple perspectives on three publicly available corpora.
 Then, we propose an adaptive multi-curricula learning framework to schedule a committee of the organized curricula.
 The framework is established upon the reinforcement learning paradigm, which automatically chooses different curricula at the evolving learning process according to the learning status of the neural dialogue generation model.
 Extensive experiments conducted on five state-of-the-art models demonstrate its learning efficiency and effectiveness with respect to 13 automatic evaluation metrics and human judgments.
\end{abstract}

\section{Introduction}

Teaching machines to converse with humans naturally and engagingly is a fundamentally interesting and challenging problem in AI research. 
Many contemporary state-of-the-art approaches~\citep{DBLP:journals/corr/SerbanSLCPCB16,DBLP:conf/naacl/LiGBGD16,DBLP:conf/acl/ZhaoZE17,DBLP:conf/aaai/ZhouHZZL18,DBLP:conf/aaai/YoungCCZBH18,DBLP:journals/acl19/hainanzhang,DBLP:conf/iclr/GuCHK19} for dialogue generation follow the data-driven paradigm: trained on a plethora of query-response pairs, the model attempts to mimic human conversations.
As a data-driven approach, the quality of generated responses in neural dialogue generation heavily depends on the training data.
As such, in order to train a robust and well-behaved model, most works obtain large-scale query-response pairs by crawling human-generated conversations from publicly available sources such as OpenSubtitles~\citep{DBLP:conf/lrec/LisonT16}.

\begin{table}[!th]
\centering
    \resizebox{0.86\columnwidth}{!}{
    \begin{tabular}{p{0.18\columnwidth}p{0.80\columnwidth}}
        \toprule
        \textbf{Context:} & This is for you. what is it? A keychain!  \\ 
        \textbf{Response:} & It's so pretty!  Can I touch it? \\ \toprule
        \textbf{Context:} & I have many pains because of the humidity. \\
        \textbf{Response:} & I also have a cough. \\ \toprule
        \textbf{Context:} & I understand now! My rakugo wasn't for anyone else. I've been doing it for myself.  \\ 
        \textbf{Response:} & Yurakutei kikuhiko. \\
        \bottomrule
    \end{tabular}
}
\caption{Examples of dialogues with different complexities in OpenSubtitles.}
\label{tbl:intro_case}
\end{table}

 However, due to the subjectivity and open-ended nature of human conversations, the complexity of training dialogues varies greatly \citep{DBLP:conf/sigdial/LisonB17}.
 Table~\ref{tbl:intro_case} shows samples drawn from OpenSubtitles~\citep{DBLP:conf/lrec/LisonT16}, which contains millions of human-human conversations converted from movie transcripts.
 The response of the third sample ``Yurakutei kikuhiko.'' looks quite strange in terms of the given query, while the first sample is clearly easier to learn.
 The noise and uneven complexity of query-response pairs impede the learning efficiency and effects of the neural dialogue generation models.
 
 Babies learn to speak by first imitating easy and exact utterances repeatedly taught by their patient parents.
 As children grow up, they learn grade by grade, from simple conversations to more complex ones.
 Inspired by such human behaviors of learning to converse, in this paper, we introduce curriculum learning to bring the neural dialogue model with easy-to-complex learning curriculum, where the model first learns from easy conversations and then gradually manages more complicated dialogues.
 Nevertheless, organizing a curriculum with increasing difficulty faces insurmountable obstacles:
 1) automatic evaluation of dialogue complexity is a non-trivial task.  
 \citet{DBLP:conf/naacl/PlataniosSNPM19} defined the difficulty for the training examples with respect to the sentence length and word rarity in neural machine translation.
 \citet{DBLP:conf/acl/SachanX16} expressed the difficulty regarding the value of the objective function.
 So far, there is no unified approach in measuring dialogue complexity.
 2) Unlike the single metric of complexity in other tasks, dialogue complexity embodies multiple aspects of attributes~\citep{DBLP:conf/naacl/SeeRKW19}---the specificity and repetitiveness of the response, the relevance between the query and the response, etc.
 As such, in this paper, we study the dialogue distributions along five aspects of attributes to gather multiple perspectives on dialogue complexity, resulting with five curricula accordingly.
 
 Conventional curriculum learning organizes the training samples into one curriculum, whereas we employ multiple curricula for dialogue learning.
 Enlightened by the phenomenon that children usually adjust the learning focus of multiple curricula dynamically in order to acquire a good mark, 
 we further propose an adaptive multi-curricula learning framework, established upon the reinforcement learning paradigm, to automatically choose different curricula at different learning stages according to the learning status of the neural dialogue generation model.
 
 Detailed analysis and experiments demonstrate that the proposed framework effectively increases the learning efficiency and gains better performances on five state-of-the-art dialogue generation models regarding three publicly available conversational corpora.
 Code for this work is available on https://github.com/hengyicai/Adaptive\_Multi-curricula\_Learning\_for\_Dialog.

\section{Curriculum Plausibility}
\label{sec:data_analysis}

 Intuitively, a well-organized curriculum should provide the model learning with easy dialogues first, and then gradually increase the curriculum difficulty.
 However, currently, there is no unified approach for dialogue complexity evaluation, where the complexity involves multiple aspects of attributes.
 In this paper, we prepare the syllabus for dialogue learning with respect to five dialogue attributes. 
 To ensure the universality and general applicability of the curriculum, we perform an in-depth investigation on three publicly available conversation corpora, PersonaChat~\citep{DBLP:conf/acl/KielaWZDUS18}, DailyDialog~\citep{DBLP:conf/ijcnlp/LiSSLCN17} and OpenSubtitles~\citep{DBLP:conf/lrec/LisonT16}, consisting of 140 248, 66 594 and 358 668 real-life conversation samples, respectively.
 
\subsection{Conversational Attributes}

\subsubsection{Specificity}
 A notorious problem for neural dialogue generation model is that the model is prone to generate generic responses.
 The most unspecific responses are easy to learn, but are short and meaningless, while the most specific responses, consisting of too many rare words, are too difficult to learn, especially at the initial learning stage.
 Following \citet{DBLP:conf/naacl/SeeRKW19}, we measure the specificity of the response in terms of each word $w$ using Normalized Inverse Document Frequency (NIDF, ranging from 0 to 1):
 \begin{equation}
    \text{NIDF}(w) = \frac{\text{IDF}(w)- \text{idf}_{min}}{\text{idf}_{max} - \text{idf}_{min}},
 \end{equation}
 where $\text{IDF}(w)=\log{\frac{N_r}{N_w}}$.
 $N_r$ is the number of responses in the training set and $N_w$ is the number of those responses that contain $w$.
 $\text{idf}_{min}$ and $\text{idf}_{max}$ are the minimum and maximum IDFs, taken over all words in the vocabulary.
 The specificity of a response $r$ is measured as the mean NIDF of the words in $r$.

\subsubsection{Repetitiveness}
 Repetitive responses are easy to generate in current auto-regressive response decoding, where response generation loops frequently,
 whereas diverse and informative responses are much more complicated for neural dialogue generation.
 We measure the repetitiveness of a response $r$ as:
\begin{equation}
    \text{REPT}(r) = \frac{\sum_{i=1}^{|r|}I({w_i \in \{w_0, \cdots, w_{i-1}\}})}{|r|},
\end{equation}
where $I(\cdot)$ is an indicator function that takes the value 1 when $w_i \in \{w_0, \cdots, w_{i-1}\}$ is true and 0 otherwise.

\subsubsection{Query-relatedness}
A conversation is considered to be coherent if the response correlates well with the given query.
For example, given a query  ``I like to paint'', the response ``What kind of things do you paint?'' is more relevant and easier to learn than another loosely-coupled response ``Do you have any pets?''.
Following previous work~\citep{DBLP:conf/nips/ZhangGGGLBD18}, we measure the query-relatedness using the cosine similarities between the query and its corresponding response in the embedding space: $\textit{cos\_sim}(\textit{sent\_emb}(c), \textit{sent\_emb}(r))$, where $c$ is the query and $r$ is the response.
The sentence embedding is computed by taking the average word embedding weighted by the smooth inverse frequency $\textit{sent\_emb}(e)=\frac{1}{|e|}\sum_{w\in{}e}\frac{0.001}{0.001 + p(w)}emb(w)$ of words~\citep{DBLP:conf/iclr/AroraLM17}, where $emb(w)$ and $p(w)$ are the embedding and the probability\footnote{
Probability is computed based on the maximum likelihood estimation on the training data.
} of word $w$ respectively.

\subsubsection{Continuity}
 A coherent response not only responds to the given query, but also triggers the next utterance.
 An interactive conversation is carried out for multiple rounds and a response in the current turn also acts as the query in the next turn.
 As such, we introduce the continuity metric, which is similar to the query-relatedness metric, to assess the continuity of a response $r$ with respect to the subsequent utterance $u$,  by measuring the cosine similarities between them. 

\subsubsection{Model Confidence}
 Despite the heuristic dialogue attributes, we further introduce the model confidence as an attribute, which distinguishes the easy-learnt samples from the under-learnt samples in terms of the model learning ability.
 A pretrained neural dialogue generation model assigns a relatively higher confidence probability for the easy-learnt samples than the under-learnt samples. 
 Inspired by \citet{DBLP:conf/nips/KumarPK10,DBLP:conf/icml/WeinshallCA18}, we employ the negative loss value of a dialogue sample under the pretrained model as the model confidence measure, indicating whether a sampled response is easy to be generated.
 Here we choose the attention-based sequence-to-sequence architecture with a cross-entropy objective as the underlying dialogue model.

\begin{figure*}[!ht]
\centering
  
\includegraphics[width=0.92\textwidth]{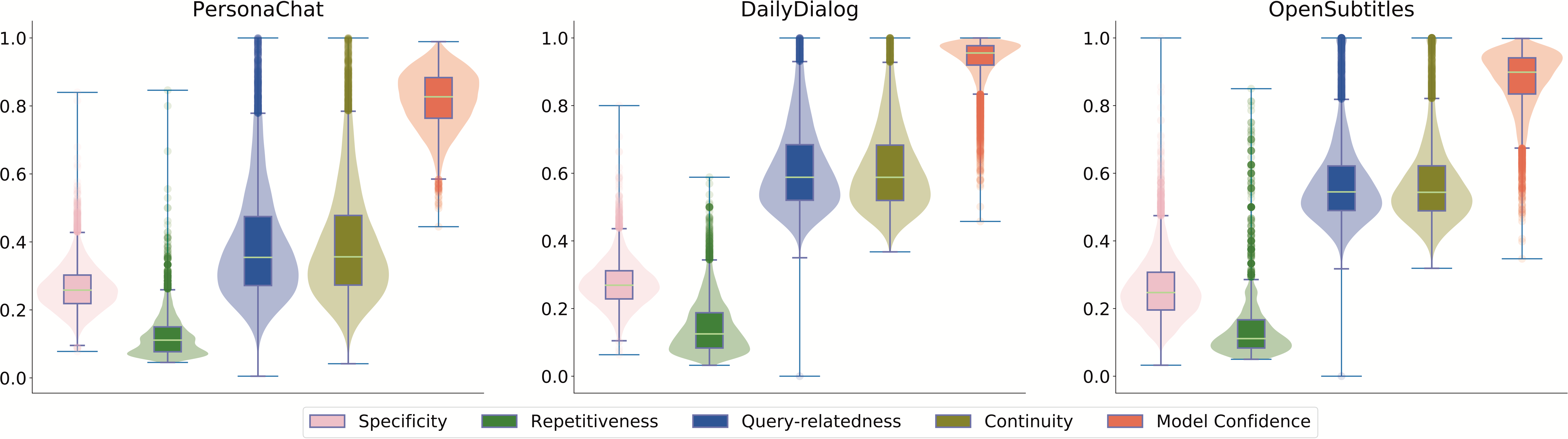}
\caption{
    Violin plot with whiskers regarding five conversation attributes in PersonaChat, DailyDialog and OpenSubtitles.
    For ease of comparison, the model confidence score is normalized by min-max normalization with the maximum and minimum confidence score on whole corpora.
}
\label{fig:box_plot}
\end{figure*}

\begin{table}[!t]
\centering
\resizebox{0.90\columnwidth}{!}{
    \begin{tabular}{lccc}
    \toprule
    \multicolumn{1}{c}{\textbf{Opponent}} & \textbf{(a)} & \textbf{(b)} & \textbf{(c)} \\
    \midrule
    Specificity vs. Repetitiveness & -0.141 & 0.065 & 0.038  \\
    Specificity vs. Query-rela.    & 0.036  & 0.127 & -0.043 \\
    Specificity vs. Model confi.   & 0.007  & -0.001& 0.001  \\
    Specificity vs. Continuity     & -0.002 & 0.093 & -0.056 \\
    Repetitiveness vs. Query-rela. & 0.002  & 0.093 & 0.011  \\
    Repetitiveness vs. Model confi.& 0.005  & 0.002 & 0.001  \\
    Repetitiveness vs. Continuity  & -0.001 & 0.059 & 0.010  \\
    Query-rela. vs. Model confi.   & -0.003 & -0.001& -0.001 \\
    Query-rela. vs. Continuity     & 0.054  & 0.150 & 0.102  \\
    Model confi. vs. Continuity    & 0.000  & -0.003& 0.000  \\
    \bottomrule
    \end{tabular}
}
\caption{Kendall $\tau$ correlations among the proposed conversational attributes on three datasets: (a) PersonaChat (b) DailyDialog and (c) OpenSubtitles.}
\label{tbl:kendall}
\end{table}

\subsection{Dialogue Analysis}
\subsubsection{Distributions among Attributes}
 The distributions of the data samples regarding the aforementioned five attributes are shown in Figure~\ref{fig:box_plot}.
 Although the attribute score distributions on three corpora are similar, they also have disparities:
 1) Outliers frequently appear among all the distributions, which exhibits the uneven dialogue complexity.
 2) In terms of query-relatedness and continuity, to our surprise, the medians of the two distributions on PersonaChat are obviously smaller than the corresponding distributions on DailyDialog and OpenSubtitles. PersonaChat is manually created by crowd-sourcing, while DailyDialog and OpenSubtitles are collected from almost real-life conversations. 
 3) With respect to the model confidence (the negative loss value), the median of PersonaChat is relatively smaller, which illustrates that it is more difficult for the neural dialogue generation model to learn from PersonaChat.

\subsubsection{Attributes Independence}
 So far, we have analyzed five dialogue attributes. 
 A question might be raised that how well the proposed attributes correlate with each other.
 To validate the correlations of these conversation attributes, we summarize the statistics of the Kendall $\tau$ correlations for each dataset in Table~\ref{tbl:kendall}.
 We find that these attributes, in general, show little correlations with each other.
 This partially validates that dialogue complexity involves multiple perspectives.

\section{Curriculum Dialogue Learning}
 We propose an adaptive multi-curricula learning framework to accelerate dialogue learning and improve the performance of the neural dialogue generation model. 

\subsection{Single Curriculum Dialogue Learning}
 We first illustrate how a dialogue generation model exploits the curriculum by taking single curriculum dialogue learning as an example, 
 where the curriculum is arranged by sorting each sample in the dialogue training set $\mathcal{D}_{train}$ according to one attribute.
 Then, at training time step $t$, a batch of training examples is sampled from the top $f(t)$ portions of the total sorted training samples, where the progressing function $f(t)$ determines the learning rate of the curriculum.
 Following~\citet{DBLP:conf/naacl/PlataniosSNPM19}, we define the progressing function $f(t)$ as $f(t)\triangleq min(1, \sqrt{t\frac{1-c_0^2}{T} + c_0^2})$, where $c_0 > 0$ is set to 0.01 and $T$ is the duration of curriculum learning.
 At the early stage of the training process, the neural dialogue generation model learns from the samples drawing from the front part of the curriculum.
 As the advance of the curriculum, the difficulty gradually increases, as more complex training examples appear.
 After training $T$ batches, each batch of training instances is drawn from the whole training set, which is same as the conventional training procedure without a curriculum.

\subsection{Adaptive Multi-curricula Learning}
 Dialogue complexity consists of multi-perspectives of attributes.
 We extend the naive single curriculum learning into the multi-curricula setting, where we provide the neural dialogue generation model with five different learning curricula, and each curriculum is prepared by ordering the training set in terms of the corresponding attribute metric accordingly.
 Scheduling multiple curricula in the same learning pace is obviously inappropriate.
 Enlightened by the phenomenon that children usually adjust the learning progress of multiple curricula dynamically in order to acquire a good mark, 
 we further introduce an adaptive multi-curricula learning framework, to automatically choose different curricula at different learning stages according to the learning status of the neural dialogue generation model.

\begin{figure}[!t]
\centering
  \includegraphics[width=0.88\columnwidth]{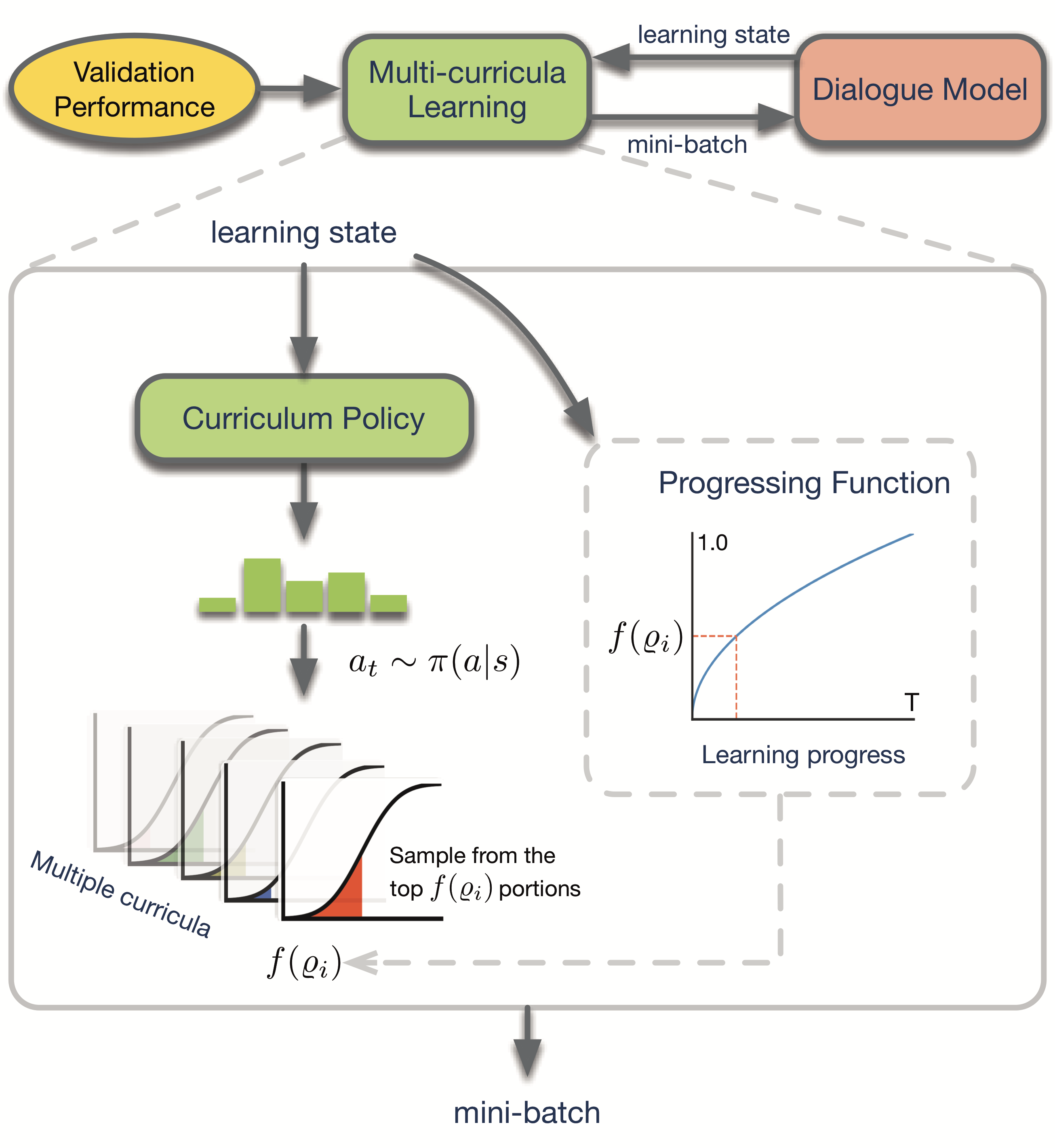}
  \caption{
    Overview of the proposed adaptive multi-curricula learning framework for neural dialogue generation.
    At training step $t$, the curriculum policy chooses one of the curricula to learn and the progressing function defines the learning progress on the selected curriculum.
  }
  \label{fig:model_arch}
\end{figure}

 The adaptive multi-curricula learning framework is established upon the reinforcement learning (RL) paradigm.
 Figure~\ref{fig:model_arch} illustrates the overall learning process.
 The multi-curricula learning scheme is scheduled according to the model's performance on the validation set, where the scheduling mechanism acts as the policy $\pi$ interacting with the dialogue model to acquire the learning status $s$.
 The reward of the multi-curricula learning mechanism $m_t$ indicates how well the current dialogue model performs.
 A positive reward is expected if a multi-curricula scheduling action $a_t$ brings improvements on the model's performance, and the current mini-batch of training samples is drawn consulting with the scheduling action $a_t$. 
 The neural dialogue generation model learns from those mini-batches, resulting with a new learning status $s_{t+1}$.
 The adaptive multi-curricula learning framework is optimized to maximize the reward.
 Such learning process loops continuously until the performance of the neural dialogue generation model converges.

 More specifically, the learning status of the dialogue model is represented as the state.
 Similar to other curriculum learning framework~\citep{DBLP:conf/icml/BengioLCW09,DBLP:conf/acl/TsvetkovFLMD16}, the learning status consists of several features, including the passed mini-batch number, the average historical training loss, the loss value on the training data, the margin value of predicted probabilities and the last validation metric values.
 To enable the proposed framework to be aware of the learning progress $\varrho_i$ regarding each attribute $i$,
 we also exploit $\varrho=\{\varrho_0, \varrho_1, \cdots, \varrho_{k-1}\}$ for state representations, where $k$ stands for the number of curricula, here $k=5$, and $\varrho_i$ can be simply measured as the learning steps on the attribute $i$.
 The multi-curricula learning framework samples a scheduling action $a_t$ per step by its policy $\Phi_\theta(a|s)$ with parameters $\theta$ to be learnt, and the scheduling action $a_t \in \{0, 1, \cdots, k-1\}$ chooses one of the curricula.
 Then, a mini-batch of dialogue instances is sampled from the top $f(\varrho_i)$ portions of the chosen curriculum.
 The dialogue model is validated every $\Gamma$ training steps and
 the curriculum policy is updated at $\Gamma$-round intervals according to a reward $m_\Gamma$.
 To accelerate the neural dialogue learning, $m_\Gamma$ is defined as the ratio of two consecutive performance deviations on a held-out validation set:  $m_\Gamma=\frac{\delta_{\Gamma}}{\delta_{\Gamma_{\text{prev}}}} - 1$. 
 The performance deviation $\delta_{\Gamma}$ is calculated in terms of $13$ automatic evaluation metrics $\{\xi_1, \xi_2, \cdots, \xi_{13}\}$ used in the experiments:
 \begin{equation}
     \delta_{\Gamma}=\sum_{i=1}^{13}(\xi^{\Gamma}_i - \xi^{\Gamma_{\text{prev}}}_i),
 \end{equation}
 where $\xi_i^{\Gamma}$ is the evaluation score of metric $i$ computed at the current validation turn and $\xi_i^{\Gamma_{\text{prev}}}$ is computed at the previous validation turn. Each score is normalized into $[0,1]$.

The curriculum policy is trained by maximizing the expected reward: $J(\theta)=\mathbb{E}_{\Phi_\theta(a|s)}[M(s,a)]$, where $M(s,a)$ is the state-action value function.
Since $M(s,a)$ is non-differentiable w.r.t. $\theta$, in this work, we use REINFORCE~\citep{DBLP:journals/ml/Williams92}, a likelihood ratio policy gradient algorithm to optimize $J(\theta)$ based on the gradient: 
\begin{equation}
\begin{aligned}
    \nabla_\theta &= \sum_{t=1}^{\Gamma}\mathbb{E}_{\Phi_\theta(a|s)}[\nabla_\theta\log{\Phi_\theta(a_t|s_t)M(s_t, a_t)}] \\
    &\approx \sum_{t=1}^{\Gamma}\nabla_\theta\log{\Phi_\theta(a_t|s_t)v_t}
\end{aligned},
\end{equation}
where $v_t$ is the sampled estimation of reward $M(s_t, a_t)$ from one episode execution of the policy $\Phi_\theta(a|s)$.
In our implementation, $v_t$ is computed as the terminal reward $m_\Gamma$.

\section{Experiments}

\begin{table*}[ht]
\centering
\resizebox{0.96\textwidth}{!}{
\centering
\begin{tabular}{p{0.05\columnwidth}|>{\rowmac}l>{\rowmac}c>{\rowmac}c>{\rowmac}c>{\rowmac}c>{\rowmac}c>{\rowmac}c>{\rowmac}c>{\rowmac}c>{\rowmac}c>{\rowmac}c>{\rowmac}c>{\rowmac}c>{\rowmac}c<{\clearrow}}
\toprule
 & Models & BLEU & Dist-1 & Dist-2 & Dist-3 & Intra-1 & Intra-2 & Intra-3 & Avg & Ext & Gre & Coh & Ent-1 & Ent-2 \\
\midrule
\multirow{10}{*}{(a)} & SEQ2SEQ & 0.316 & 0.3967 & 2.190 & 5.026 & 77.24 & 87.00 & 90.73 & 58.85 & 47.22 & 65.91 & 62.87 & 6.674 & 10.368 \\
\setrow{\bfseries} 
 &SEQ2SEQ (${\blacktriangle}$) & 0.352 & 0.5406 & 3.537 & 8.343  & 82.74 & 92.28 & 95.47 & 62.20 & 47.57 & 67.07 & 66.87 & 6.875 & 10.723   \\ \cline{2-15} 
 &CVAE & 0.290 & 0.5330 & 3.228 & 7.715  & 85.30 & 94.69 & 96.73 & 61.99 & 47.12 & 66.68 & 65.27 & 6.900 & 10.738   \\ 
\setrow{\bfseries} 
 &CVAE (${\blacktriangle}$)    & 0.321 & 0.6530 & 4.572 & 11.326 & 89.39 & 96.90 & 98.28 & 63.08 & 47.37 & 67.09 & 66.81 & 6.973 & 10.866   \\ \cline{2-15}
 &Transformer & 0.195 & 0.7667 & 3.264 & 6.262  & 83.11 & 93.82 & 96.48 & 59.53 & 44.99 & 65.57 & 62.48 & 7.169 & 11.232 \\
\setrow{\bfseries} 
 &Transformer (${\blacktriangle}$) & 0.329 & 0.8468 & 4.291 & 8.829  & 89.39 & 97.92 & 99.29 & 62.33 & 46.24 & 66.54 & 65.35 & 7.183 & 11.331   \\ \cline{2-15}
 &HRED & 0.272 & 0.8109 & 4.217 & 8.948  & 84.25 & 95.20 & 97.19 & 60.63 & 45.91 & 66.33 & 63.53 & 7.082 & 11.149  \\ 
\setrow{\bfseries} 
 &HRED (${\blacktriangle}$)        & 0.308 & 0.8926 & 5.332 & 12.281 & 91.45 & 97.89 & 98.93 & 62.25 & 46.53 & 66.53 & 65.22 & 7.156 & 11.274 \\ \cline{2-15}
 &DialogWAE & 0.124 & 0.9594 & 5.153 & 11.483 & 94.35 & 98.04 & 98.54 & 58.98 & 43.53 & 63.66 & 60.93 & 7.424 & 11.696  \\ 
\setrow{\bfseries} 
 &DialogWAE (${\blacktriangle}$)   & 0.171 & 1.1388 & 6.890 & 15.842 & 96.65 & 99.41 & 99.68 & 63.81 & 45.90 & 65.63 & 65.63 & 7.462 & 11.845   \\ 
\bottomrule
\bottomrule
\multirow{10}{*}{(b)} & SEQ2SEQ & 0.399 & 1.542 & 9.701 & 22.005 & 91.10   & 96.97   & 98.15  & 67.50   & 47.41   & 68.45  & 68.39 & 6.933 & 10.921 \\ 
\setrow{\bfseries} 
 &SEQ2SEQ (${\blacktriangle}$)     & 0.617 & 1.846 & 11.665 & 25.918 & 93.28  & 98.16   & 99.00  & 67.75   & 47.57   & 68.91  & 68.94 & 7.041 & 11.164 \\ \cline{2-15}
 &CVAE & 0.406 & 1.615 & 11.187 & 26.588 & 90.56 & 97.48 & 98.70 & 67.76 & 46.82 & 68.90 & 67.77 & 7.124 & 11.308  \\ 
\setrow{\bfseries} 
 &CVAE (${\blacktriangle}$)        & 0.691  & 1.890 & 13.125 & 30.793 & 94.48 & 98.88 & 99.47 & 67.81 & 47.36 & 69.00 & 68.00 & 7.139 & 11.453  \\ \cline{2-15}
 &Transformer & 0.412  & 2.617 & 13.212 & 25.175 & 90.50 & 96.53 & 97.92 & 65.82 & 46.01 & 67.86 & 66.03 & 7.192 & 11.309  \\ 
\setrow{\bfseries}
 &Transformer (${\blacktriangle}$) & 0.8063 & 2.917 & 15.509 & 30.954 & 94.38 & 98.59 & 99.26 & 66.52 & 46.79 & 68.40 & 66.65 & 7.307 & 11.651 \\ \cline{2-15}
 &HRED & 0.1746 & 2.323 & 11.563 & 22.471 & 94.01 & 98.45 & 99.30 & 65.09 & 45.91 & 67.49 & 65.09 & \textbf{7.141} & 11.331 \\ 
\setrow{\bfseries}
 &HRED (${\blacktriangle}$)        & 0.3834 & 2.448 & 12.880 & 26.355 & 94.18 & 98.65 & 99.36 & 65.37 & 46.43 & 68.14 & 65.22 & \normalfont{7.058} & 11.341 \\ \cline{2-15}
 &DialogWAE & 0.0303 & 2.244 & 12.340 & 26.109 & \textbf{92.98} & 98.02 & 98.78 & 64.19 & 42.03 & 65.52 & 64.31 & 7.420 & 11.954 \\ 
\setrow{\bfseries}
 &DialogWAE (${\blacktriangle}$)   & 0.0814 & 2.654 & 16.311 & 36.591 & \normalfont{92.79} & 98.73 & 99.53 & 65.27 & 43.41 & 66.60 & 65.62 & 7.539 & 12.106 \\
\bottomrule
\bottomrule
\multirow{10}{*}{(c)} & SEQ2SEQ & 0.140 & 0.3053 & 2.472 & 6.377 & 95.94 & 97.37 & 98.34 & 54.71 & 49.03 & 62.87 & 59.09 & 6.226 & 9.516 \\ 
\setrow{\bfseries} 
 &SEQ2SEQ (${\blacktriangle}$)     & 0.172 & 0.4870 & 4.514 & 12.319 & 96.67 & 98.16 & 98.76 & 55.87 & 49.13 & 63.78 & 62.65 & 6.353 & 10.236 \\ \cline{2-15}
 &CVAE & \textbf{0.0522} & 0.3028 & 2.614 & 7.574 & 95.12 & 97.32 & 98.19 & 56.17 & 47.70 & 63.10 & 58.85 & 6.156 & 9.460  \\ 
\setrow{\bfseries} 
 &CVAE (${\blacktriangle}$)        & \normalfont{0.0429} & 0.4061 & 3.928 & 12.676 & 96.11 & 98.09 & 98.99 & 57.06 & 47.85 & 63.44 & 60.82 & 6.463 & 10.442  \\ \cline{2-15}
 &Transformer             & \textbf{0.072}  & 0.3883 & 1.737 & 3.503 & 95.38 & 97.13 & 98.18 & 55.10 & 48.16 & 62.69 & 57.45 & 6.661 & 10.362 \\ 
\setrow{\bfseries}
 &Transformer (${\blacktriangle}$) & \normalfont{0.050}  & 0.5655 & 3.079 & 7.005 & 97.15 & 98.39 & 99.11 & 55.63 & 48.17 & 63.16 & 59.19 & 6.666 & 10.715 \\ \cline{2-15}
 &HRED                    & 0.0498 & 0.3311 & 1.900 & 4.465 & 95.34 & 97.38 & 98.15 & 55.41 & 48.34 & 62.79 & 58.92 & 6.346 & 9.715  \\ 
\setrow{\bfseries}
 &HRED (${\blacktriangle}$)        & 0.0795 & 0.6982 & 4.224 & 9.933 & 97.43 & 98.68 & 99.20 & 55.89 & 48.64 & 63.53 & 59.55 & 6.510 & 10.409 \\ \cline{2-15}
 &DialogWAE               & 0.0038 & 0.4808 & 3.870 & 11.856 & 86.91 & 93.88 & 97.93 & 51.59 & 43.40 & 56.23 & 51.96 & 5.633 & 8.559 \\  
\setrow{\bfseries}
 &DialogWAE (${\blacktriangle}$)   & 0.0352 & 0.7360 & 6.549 & 18.881 & 94.92 & 97.10 & 98.14 & 54.73 & 47.84 & 63.52 & 58.81 & 6.7859 & 11.187 \\
\bottomrule
\end{tabular}
}
\caption{Automatic evaluation results (\%) on the test set of three datasets: (a) PersonaChat, (b) DailyDialog and (c) OpenSubtitles. 
``${\blacktriangle}$'' denotes that the model is trained using our proposed framework.
{The metrics Average, Extrema, Greedy and Coherence are abbreviated as Avg, Ext, Gre and Coh, respectively.}
The best results in each group are highlighted with \textbf{bold}.}
\label{tbl:main_res}
\end{table*}

\subsection{Experiment Settings}
We perform experiments using the following state-of-the-art models:
(\romannum{1}) SEQ2SEQ: a sequence-to-sequence model with attention mechanisms~\citep{Bahdanau2014NeuralMT},
(\romannum{2}) CVAE:  a conditional variational auto-encoder model with KL-annealing and a BOW loss~\citep{DBLP:conf/acl/ZhaoZE17},
(\romannum{3}) Transformer:  an encoder-decoder architecture relying solely on attention mechanisms~\citep{DBLP:conf/nips/VaswaniSPUJGKP17},
(\romannum{4}) HRED:  a generalized sequence-to-sequence model with the hierarchical RNN encoder~\citep{DBLP:conf/aaai/SerbanSBCP16},
(\romannum{5}) DialogWAE:  a conditional Wasserstein auto-encoder, which models the distribution of data by training a GAN within the latent variable space~\citep{DBLP:conf/iclr/GuCHK19}.
We adopt several standard metrics widely used in existing works to measure the performance of dialogue generation models,
including BLEU~\citep{DBLP:conf/acl/PapineniRWZ02}, embedding-based metrics {(Average, Extrema, Greedy and Coherence)}~\citep{DBLP:conf/emnlp/LiuLSNCP16,DBLP:conf/emnlp/XuDKR18}, entropy-based metrics {(Ent-\{1,2\})}~\citep{DBLP:journals/corr/SerbanSLCPCB16} and distinct metrics {(Dist-\{1,2,3\} and Intra-\{1,2,3\})}~\citep{DBLP:conf/naacl/LiGBGD16,DBLP:conf/iclr/GuCHK19}.

\subsection{Implementation and Reproducibility}
Our experiments are performed using ParlAI~\citep{miller-etal-2017-parlai}.
Regarding model implementations, we employ a 2-layer bidirectional LSTM as the encoder and a unidirectional one as the decoder for the SEQ2SEQ and CVAE.
The hidden size is set to 512, and the latent size is set to 64 for CVAE.
For the Transformer, the hidden size, attention heads and number of hidden layers are set to 512, 8 and 6, respectively.
In terms of HRED and DialogWAE, the utterance encoder is a bidirectional GRU with 512 hidden units in each direction.
The context encoder and decoder are both GRUs with 512 hidden units.
Regarding the curriculum length $T$, we set its value in the following manner:
we train the baseline model using the vanilla training procedure and compute the number of training steps it takes to reach approximately 110\% of its final loss value.
We then set $T$ to this value.
Each model is trained using two protocols: 
the vanilla training procedure without using any curriculum and our proposed adaptive multi-curricula learning procedure, keeping other configurations the same.

\subsection{Overall Performance and Human Evaluation}
The automatic evaluation results of our proposed multi-curricula learning framework and the comparison models are listed in Table~\ref{tbl:main_res}.
Compared with the vanilla training procedure, our curriculum learning framework
1) brings solid improvements for all the five dialogue models regarding almost all the evaluation metrics,
2) achieves competitive performance across three datasets, affirming the superiority and general applicability of our proposed framework.
We also notice that the relative improvements of Distinct on OpenSubtitles are much larger (up to 122.46\%) than the other two experiment datasets.
We conjecture that the OpenSubtitles, with extremely uneven-complexity dialogue samples, benefits more from the multi-curricula learning paradigm.

We conduct a human evaluation to validate the effectiveness of the proposed multi-curricula learning framework.
We employ the DailyDialog as the evaluation corpus since it is closer to our daily conversations and easier for humans to make the judgment.
We randomly sampled 100 cases from the test set and compared the generated responses of the models trained with the vanilla learning procedure and the multi-curricula learning framework.
Three annotators, who have no knowledge about which system the response is from, 
are then required to evaluate among win (response$_1$ is better), loss (response$_2$ is better) and tie (they are equally good or bad) independently, considering four aspects: coherence, logical consistency, fluency and diversity. 
Cases with different rating results are counted as ``tie''.
Table~\ref{tbl:human_eval} reveals the results of the subjective evaluation.
We observe that our multi-curricula learning framework outperforms the vanilla training method on all the five dialogue models and the kappa scores indicate that the annotators came to a fair agreement in the judgment.
We checked the cases on which the vanilla training method loses to our multi-curricula learning method and found that the vanilla training method usually leads to irrelevant, generic and repetitive responses, while our method effectively alleviates such defects.

\begin{table}[!htp]
\centering
\resizebox{0.68\columnwidth}{!}{
\begin{tabular}{@{}lcccc@{}}
\toprule
\makecell{\textbf{Our method} \textit{vs.}\\ \textbf{Vanilla training}}   & \textbf{Win} & \textbf{Loss} & \textbf{Tie} & \textbf{Kappa}  \\ \midrule
\textbf{SEQ2SEQ}      & 52\%  & 7\%  &  41\% & 0.5137  \\  
\textbf{CVAE}         & 42\%  & 5\%  & 53\%  & 0.4633  \\  
\textbf{Transformer}  & 41\%  &  14\% & 45\%  & 0.4469  \\   
\textbf{HRED}         & 46\%  & 12\%  & 42\%  & 0.5940  \\   
\textbf{DialogWAE}    & 42\%  & 9\%  &  49\% & 0.5852  \\  \bottomrule 
\end{tabular}
}
\caption{The results of human evaluation on DailyDialog.}
\label{tbl:human_eval}
\end{table}

\begin{table*}[!htp]
\centering
\resizebox{0.92\textwidth}{!}{
\centering
\begin{tabular}{>{\rowmac}l>{\rowmac}c>{\rowmac}c>{\rowmac}c>{\rowmac}c>{\rowmac}c>{\rowmac}c>{\rowmac}c>{\rowmac}c>{\rowmac}c>{\rowmac}c>{\rowmac}c>{\rowmac}c>{\rowmac}c<{\clearrow}}
\toprule
Curriculum by       & BLEU & Dist-1 & Dist-2 & Dist-3 & Intra-1 & Intra-2 & Intra-3 & Avg & Ext & Gre & Coh & Ent-1 & Ent-2 \\
\midrule
\textbf{None}             & 0.3599  & 1.5777 & 10.103 & 23.149 & 90.99 & 96.40 & 97.63 & 67.49 & 47.15 & 68.47 & 68.44 & 7.065 & 11.077    \\
\midrule
\textbf{Specificity}         & 0.6866  & 1.6957 & 11.828 & \textbf{28.883} & 91.63 & 97.77 & 98.83 & 67.64 & 47.02 & 68.77 & 68.25 & 7.091 & 11.281    \\
\textbf{Repetitiveness}          & 0.5994  & 1.6581 & 10.590 & 23.209 & 91.63 & 97.20 & 98.25 & 67.72 & 47.45 & 68.53 & 67.70 & 7.038 & 11.085    \\
\textbf{Query-relatedness}       & 0.5745  & 1.7125 & 11.461 & 26.625 & 90.27 & 96.64 & 97.99 & 67.79 & 46.90 & 68.56 & 69.33 & 7.070 & 11.121    \\
\textbf{Continuity}          & 0.5662  & 1.7496 & 11.370 & 25.710 & 91.83 & 97.56 & 98.58 & 67.55 & 47.37 & 68.66 & 67.89 & 7.036 & 11.105    \\
\textbf{Model confidence}          & 0.5917  & 1.7264 & 11.365 & 26.145 & 91.47 & 97.31 & 98.40 & 67.44 & 46.85 & 68.57 & 68.17 & 7.092 & 11.172    \\
\midrule
\setrow{\bfseries}
Multi-curricula             & 0.7057  & 1.9612 & 12.705 & \normalfont{28.759} & 92.93 & 98.13 & 98.89 & 67.82 & 47.46 & 68.81 & 69.35 & 7.101 & 11.289    \\
\bottomrule

\end{tabular}
}
\caption{Ablation test (\%) for the proposed five curriculum learning attributes on the validation set of the DailyDialog dataset with the SEQ2SEQ model.}
\label{tbl:ablation_test}
\end{table*}

\begin{table*}[!htp]
\centering
\resizebox{0.92\textwidth}{!}{
\begin{tabular}{>{\rowmac}l>{\rowmac}c>{\rowmac}c>{\rowmac}c>{\rowmac}c>{\rowmac}c>{\rowmac}c>{\rowmac}c>{\rowmac}c>{\rowmac}c>{\rowmac}c>{\rowmac}c>{\rowmac}c>{\rowmac}c<{\clearrow}}
\toprule
  & BLEU & Dist-1 & Dist-2 & Dist-3 & Intra-1 & Intra-2 & Intra-3 & Avg & Ext & Gre & Coh & Ent-1 & Ent-2 \\
\midrule
\setrow{\bfseries}
Full method        & 0.617 & 1.846 & 11.665 & 25.918 & 93.28  & 98.16  & 99.00  & 67.75   & 47.57   & 68.91  & 68.94 & 7.041 & 11.164  \\
(a) \textit{with random policy}   & 0.565 & 1.696 & 10.563 & 23.383 & 92.57 & 97.22 & 98.85  & 65.64   & 46.96   & 67.96  & 66.64 & 6.939 & 10.953  \\
(b) \textit{with anti-curriculum} & 0.354 & 1.491 & 9.6950 & 21.978 & 90.48 & 96.69  & 97.72  & 67.09   & 46.87   & 68.19 & 66.85 & 6.973 & 11.025  \\
\bottomrule
\end{tabular}
}
\caption{Ablation test (\%) of the curriculum learning framework on the DailyDialog dataset with SEQ2SEQ.}
\label{tbl:ablation_policy}
\end{table*}

\begin{figure}[!th]
\centering
  \includegraphics[width=0.90\columnwidth]{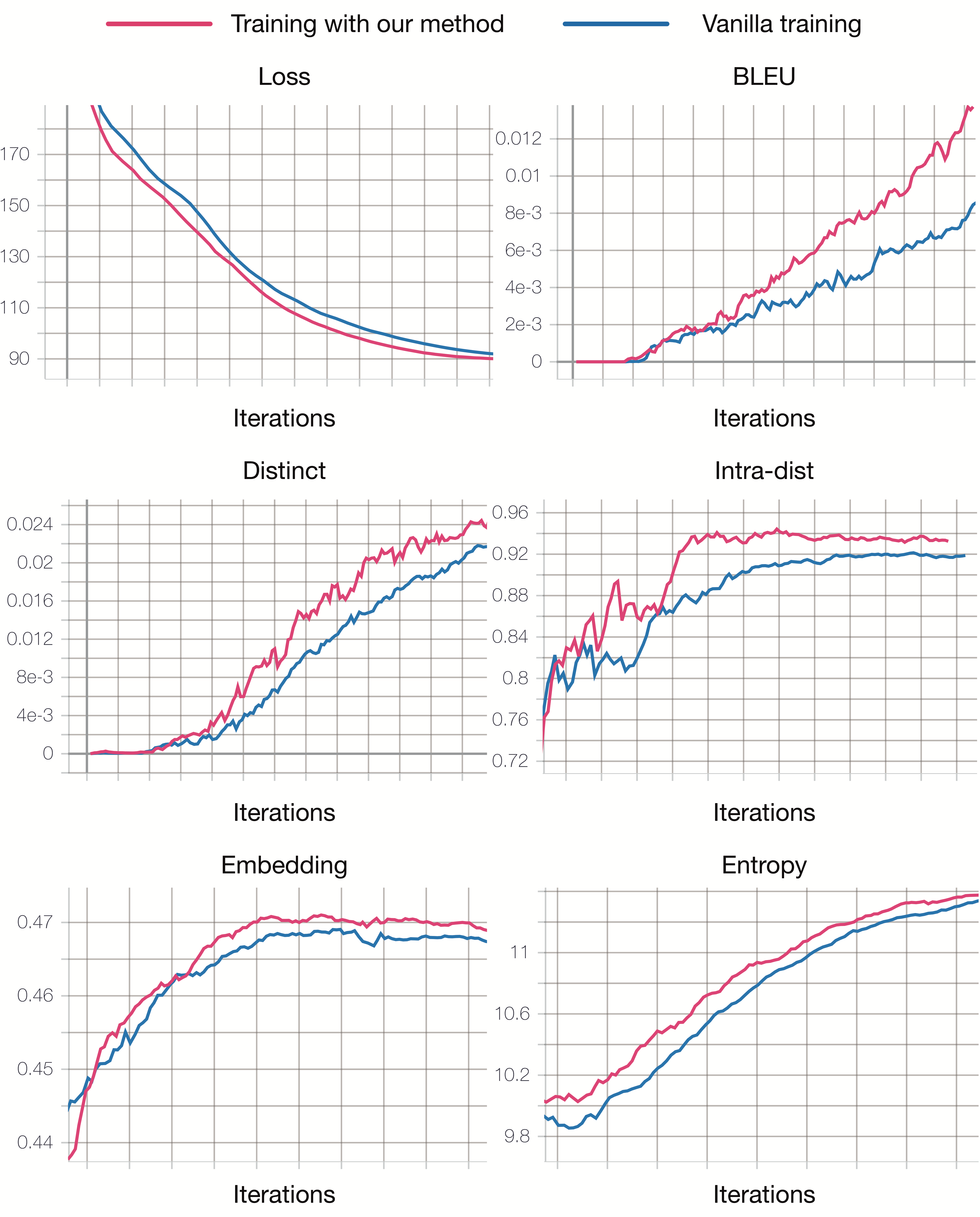}
  \caption{
  Comparison of the vanilla training and adaptive multi-curricula learning for six evaluation metrics with SEQ2SEQ on the validation set of  DailyDialog. Dist-1, Intra-1, Embedding Extrema and Ent-2 are denoted as ``Distinct'', ``Intra-dist'', ``Embedding'' and ``Entropy'', respectively.
  }
  \label{fig:valid_line_plot}
\end{figure}

\begin{figure}[!htp]
\centering
  \includegraphics[width=0.86\columnwidth]{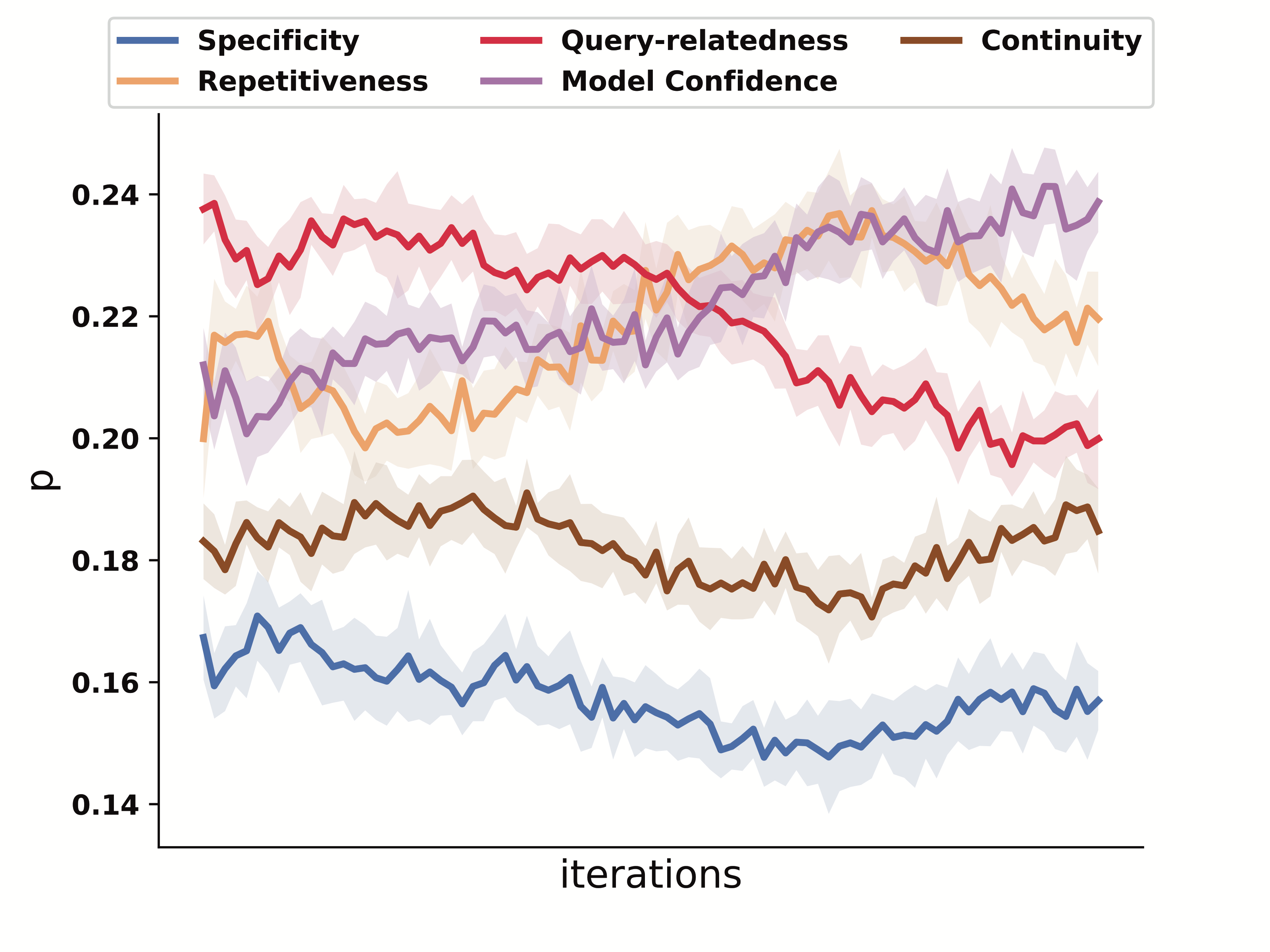}
  \caption{
    Trajectory of the curriculum learning action distribution $\pi(a_t|s_t)$ during the learning process.
    The bold line shows the mean and the shaded region shows the standard deviation.
  }
  \label{fig:subtask_p_during_training}
\end{figure}

\begin{table}[!htp]
\centering
\resizebox{0.95\columnwidth}{!}{
    \begin{tabular}{p{0.18\columnwidth}p{0.81\columnwidth}}
        \toprule
        \multicolumn{2}{c}{\textbf{Examples with top learning frequencies}} \\ \midrule
        \textbf{Context:} & Have you ever been to America?  \\ 
        \textbf{Response:} & No, but I'd really like to. \\ 
        \hline
        \textbf{Context:} & Can i ask you a question? \\
        \textbf{Response:} & Sure, what do you want to know? \\
        \hline
        \textbf{Context:} & A new film is on. Have you ever seen it? \\
        \textbf{Response:} & What kind of movie is it? \\
        \toprule
        \toprule
        \multicolumn{2}{c}{\textbf{Examples with tail learning frequencies}} \\ \midrule
        \textbf{Context:} & Ma'am?  \\ 
        \textbf{Response:} & May I have a magazine or something? \\ 
        \hline
        \textbf{Context:} & What took you so long? \\
        \textbf{Response:} & Gross. Diarrhea? \\
        \hline
        \textbf{Context:} & I need more water! \\
        \textbf{Response:} & oooo... gross... You're sweating all over the place. \\
        \bottomrule
    \end{tabular}
}
\caption{Examples with top and tail learning frequencies of the SEQ2SEQ model on DailyDialog, using the proposed learning framework.}
\label{tbl:sample_count}
\end{table}

\subsection{Model Analysis}
\subsubsection{Single \textit{vs} Multi-curricula} To further glean the insights regarding the effects of the five conversational attributes on the proposed learning framework,
we conduct the ablation test using the SEQ2SEQ model by only exploiting a single attribute during the curriculum learning.
Table~\ref{tbl:ablation_test} reports the ablation test results on the DailyDialog.
We observe that the curriculum learning leads to consistent performance improvements, even with one single conversational attribute.
When applying the multi-curricula learning method to the model, we observe the nearly best performance.
\noindent
\subsubsection{Effects of Adaptive Multi-curricula Learning}
Adaptive multi-curricula learning enables the model to choose different curricula at different learning stages according to the learning status of the underlying model.
As shown in Table~\ref{tbl:ablation_policy}, we notice the performance drops when replacing the RL-based curriculum policy with the random policy,
indicating that choosing different curricula according to the learning status of the model benefits the model training.
When training the model with anti-curriculum learning, i.e., feeding examples to the model in the complex-to-easy manner, we also observe consistent performance decreases, affirming the effectiveness of the easy-to-complex learning manner.
\noindent
\subsubsection{Learning Efficiency}  Figure~\ref{fig:valid_line_plot} shows comparative results when training the SEQ2SEQ model on DailyDialog with different training protocols.
As shown in Figure~\ref{fig:valid_line_plot},
our training method accelerates the learning effectively and consistently outperforms the baseline by a large margin in most cases.
\noindent
\subsubsection{Multi-curricula Learning Route}
 To glean insights on how the proposed adaptive multi-curricula learning framework performs, we present the choosing curriculum distributions $\pi(a_t|s_t)$ during the model learning in Figure~\ref{fig:subtask_p_during_training}.
 We notice that the model focuses more on the curriculum of  ``query-relatedness'' at the initial learning stage.
 As the learning proceeds, the model gradually turns its attention to other curricula.
 At the final stage, the model pays more attention to the ``model confidence'' curriculum.
 Such dynamic learning route is quite similar to the human learning behavior.
\noindent
\subsubsection{Examples with Different Learning Frequencies}
 As shown in Table~\ref{tbl:sample_count}, the most frequently learnt examples are comprehensively far better than those seldom learnt examples, which exhibits the effectiveness of the adaptive multi-curricula learning framework. 

\section{Related Work}

\textit{Neural dialogue generation.}\quad
Neural generation models for dialogue, despite their ubiquity in current research, are still far from the real-world applications.
Previous approaches enhancing neural dialogue generation models mainly focus on the learning systems by incorporating extra information to the dialogue models such as relevant dialogue history~\citep{DBLP:journals/acl19/hainanzhang}, topics~\citep{DBLP:conf/aaai/XingWWLHZM17}, emotions~\citep{DBLP:conf/aaai/ZhouHZZL18}, out-sourcing knowledge~\citep{DBLP:conf/aaai/YoungCCZBH18} or exemplars~\citep{DBLP:conf/acl/ContractorKJP18}.
Latent variables~\citep{DBLP:journals/corr/SerbanSLCPCB16,DBLP:conf/acl/ZhaoZE17} also benefit the model with more diverse response generations.
In contrast with the previous researches, which pay most attention to the underlying dialogue models, in this work, 
we concentrate on the dialogue learning process and investigate how the performance of existing dialogue models can be improved on the conversation corpora with varying levels of complexity, by simply adapting the training protocols.
\citet{DBLP:conf/acl/Csaky2019} attributed the generic/uninteresting responses to the high-entropy utterances in the training set and proposed to improve dataset quality through data filtering.
Though straightforward, the filtering threshold need be carefully chosen to prevent the data size decreasing too much.
\citet{DBLP:conf/sigdial/LisonB17,DBLP:conf/ijcai/ShangFPFZY18} proposed to investigate instance weighting into dialogue systems.
However, it is difficult to accurately define the ``weight'' of an example in conversation systems, since the dialogue data is of high diversity and complexity.
Our proposed adaptive multi-curricula learning framework, concentrating on different curricula at evolving learning process according to the learning status of the underlying model, enables dialogue systems gradually proceed from easy to more complex samples in training and thus efficiently improves the response quality.

\noindent
\textit{Curriculum learning in NLP.}\quad
\citet{DBLP:conf/icml/BengioLCW09} examined curriculum learning and demonstrated empirically that such curriculum approaches indeed help decrease training times and sometimes even improve generalization.
\citet{DBLP:conf/aaai/JiangMZSH15} managed curriculum learning as an optimization problem.
Curriculum learning has also been applied to many NLP tasks.
To name a few, \citet{DBLP:conf/acl/SachanX16} applied self-paced learning for neural question answering.
 \citet{DBLP:conf/ijcai/LiuH0018} proposed a curriculum learning based natural answer generation framework, dealing with low-quality QA-pairs first and then gradually learn more complete answers.
\citet{DBLP:journals/acl/TayYi2019} proposed curriculum pointer-generator networks for reading comprehension over long narratives.
\citet{DBLP:conf/naacl/PlataniosSNPM19} applied curriculum learning for neural machine translation (NMT), aiming to reduce the need for specialized training heuristics and boost the performance of existing NMT systems.
In our work, instead of organizing the curriculum only from a single aspect, we provide an adaptive multi-curricula dialogue learning framework, grounding our analysis on five conversation attributes regarding the dialogue complexity.

\section{Conclusion}
In this paper, we propose an adaptive multi-curricula dialogue learning framework, to enable the dialogue models to gradually proceed from easy samples to more complex ones in training.
We first define and analyze five conversational attributes regarding the complexity and easiness of dialogue samples, and then present an adaptive multi-curricula learning framework, which chooses different curricula at different training stages according to the learning status of the model.
Extensive experiments conducted on three large-scale datasets and five state-of-the-art conversation models show that our proposed learning framework is able to boost the performance of existing dialogue systems.

\section*{Acknowledgments}

This work is supported by the National Natural Science Foundation of China-Joint Fund for Basic Research of General Technology under Grant U1836111 and U1736106.
Hongshen Chen and Yonghao Song are the corresponding authors.

\bibliographystyle{aaai}
\fontsize{9.0pt}{10.0pt}
\bibliography{3925-aaai}

\begin{thebibliography}{}

\bibitem[\protect\citeauthoryear{Arora, Liang, and
  Ma}{2017}]{DBLP:conf/iclr/AroraLM17}
Arora, S.; Liang, Y.; and Ma, T.
\newblock 2017.
\newblock A simple but tough-to-beat baseline for sentence embeddings.
\newblock In {\em ICLR}.

\bibitem[\protect\citeauthoryear{Bahdanau, Cho, and
  Bengio}{2015}]{Bahdanau2014NeuralMT}
Bahdanau, D.; Cho, K.; and Bengio, Y.
\newblock 2015.
\newblock Neural machine translation by jointly learning to align and
  translate.
\newblock In {\em ICLR}.

\bibitem[\protect\citeauthoryear{Bengio \bgroup et al\mbox.\egroup
  }{2009}]{DBLP:conf/icml/BengioLCW09}
Bengio, Y.; Louradour, J.; Collobert, R.; and Weston, J.
\newblock 2009.
\newblock Curriculum learning.
\newblock In {\em ICML}.

\bibitem[\protect\citeauthoryear{Csaky, Purgai, and
  Recski}{2019}]{DBLP:conf/acl/Csaky2019}
Csaky, R.; Purgai, P.; and Recski, G.
\newblock 2019.
\newblock Improving neural conversational models with entropy-based data
  filtering.
\newblock In {\em ACL}.

\bibitem[\protect\citeauthoryear{Gu \bgroup et al\mbox.\egroup
  }{2019}]{DBLP:conf/iclr/GuCHK19}
Gu, X.; Cho, K.; Ha, J.; and Kim, S.
\newblock 2019.
\newblock Dialogwae: Multimodal response generation with conditional
  wasserstein auto-encoder.
\newblock In {\em ICLR}.

\bibitem[\protect\citeauthoryear{Jiang \bgroup et al\mbox.\egroup
  }{2015}]{DBLP:conf/aaai/JiangMZSH15}
Jiang, L.; Meng, D.; Zhao, Q.; Shan, S.; and Hauptmann, A.~G.
\newblock 2015.
\newblock Self-paced curriculum learning.
\newblock In {\em AAAI}.

\bibitem[\protect\citeauthoryear{Kumar, Packer, and
  Koller}{2010}]{DBLP:conf/nips/KumarPK10}
Kumar, M.~P.; Packer, B.; and Koller, D.
\newblock 2010.
\newblock Self-paced learning for latent variable models.
\newblock In {\em NIPS}.

\bibitem[\protect\citeauthoryear{Li \bgroup et al\mbox.\egroup
  }{2016}]{DBLP:conf/naacl/LiGBGD16}
Li, J.; Galley, M.; Brockett, C.; Gao, J.; and Dolan, B.
\newblock 2016.
\newblock A diversity-promoting objective function for neural conversation
  models.
\newblock In {\em NAACL-HLT}.

\bibitem[\protect\citeauthoryear{Li \bgroup et al\mbox.\egroup
  }{2017}]{DBLP:conf/ijcnlp/LiSSLCN17}
Li, Y.; Su, H.; Shen, X.; Li, W.; Cao, Z.; and Niu, S.
\newblock 2017.
\newblock Dailydialog: {A} manually labelled multi-turn dialogue dataset.
\newblock In {\em IJCNLP}.

\bibitem[\protect\citeauthoryear{Lison and
  Bibauw}{2017}]{DBLP:conf/sigdial/LisonB17}
Lison, P., and Bibauw, S.
\newblock 2017.
\newblock Not all dialogues are created equal: Instance weighting for neural
  conversational models.
\newblock In {\em SIGDIAL}.

\bibitem[\protect\citeauthoryear{Lison and
  Tiedemann}{2016}]{DBLP:conf/lrec/LisonT16}
Lison, P., and Tiedemann, J.
\newblock 2016.
\newblock Opensubtitles2016: Extracting large parallel corpora from movie and
  {TV} subtitles.
\newblock In {\em LREC}.

\bibitem[\protect\citeauthoryear{Liu \bgroup et al\mbox.\egroup
  }{2016}]{DBLP:conf/emnlp/LiuLSNCP16}
Liu, C.; Lowe, R.; Serban, I.; Noseworthy, M.; Charlin, L.; and Pineau, J.
\newblock 2016.
\newblock How {NOT} to evaluate your dialogue system: An empirical study of
  unsupervised evaluation metrics for dialogue response generation.
\newblock In {\em EMNLP}.

\bibitem[\protect\citeauthoryear{Liu \bgroup et al\mbox.\egroup
  }{2018}]{DBLP:conf/ijcai/LiuH0018}
Liu, C.; He, S.; Liu, K.; and Zhao, J.
\newblock 2018.
\newblock Curriculum learning for natural answer generation.
\newblock In {\em IJCAI}.

\bibitem[\protect\citeauthoryear{Miller \bgroup et al\mbox.\egroup
  }{2017}]{miller-etal-2017-parlai}
Miller, A.; Feng, W.; Batra, D.; Bordes, A.; Fisch, A.; Lu, J.; Parikh, D.; and
  Weston, J.
\newblock 2017.
\newblock {P}arl{AI}: A dialog research software platform.
\newblock In {\em EMNLP}.

\bibitem[\protect\citeauthoryear{Pandey \bgroup et al\mbox.\egroup
  }{2018}]{DBLP:conf/acl/ContractorKJP18}
Pandey, G.; Contractor, D.; Kumar, V.; and Joshi, S.
\newblock 2018.
\newblock Exemplar encoder-decoder for neural conversation generation.
\newblock In {\em ACL}.

\bibitem[\protect\citeauthoryear{Papineni \bgroup et al\mbox.\egroup
  }{2002}]{DBLP:conf/acl/PapineniRWZ02}
Papineni, K.; Roukos, S.; Ward, T.; and Zhu, W.
\newblock 2002.
\newblock Bleu: a method for automatic evaluation of machine translation.
\newblock In {\em ACL}.

\bibitem[\protect\citeauthoryear{Platanios \bgroup et al\mbox.\egroup
  }{2019}]{DBLP:conf/naacl/PlataniosSNPM19}
Platanios, E.~A.; Stretcu, O.; Neubig, G.; P{\'{o}}czos, B.; and Mitchell,
  T.~M.
\newblock 2019.
\newblock Competence-based curriculum learning for neural machine translation.
\newblock In {\em NAACL-HLT}.

\bibitem[\protect\citeauthoryear{Sachan and
  Xing}{2016}]{DBLP:conf/acl/SachanX16}
Sachan, M., and Xing, E.~P.
\newblock 2016.
\newblock Easy questions first? {A} case study on curriculum learning for
  question answering.
\newblock In {\em ACL}.

\bibitem[\protect\citeauthoryear{See \bgroup et al\mbox.\egroup
  }{2019}]{DBLP:conf/naacl/SeeRKW19}
See, A.; Roller, S.; Kiela, D.; and Weston, J.
\newblock 2019.
\newblock What makes a good conversation? how controllable attributes affect
  human judgments.
\newblock In {\em NAACL-HLT}.

\bibitem[\protect\citeauthoryear{Serban \bgroup et al\mbox.\egroup
  }{2016}]{DBLP:conf/aaai/SerbanSBCP16}
Serban, I.~V.; Sordoni, A.; Bengio, Y.; Courville, A.~C.; and Pineau, J.
\newblock 2016.
\newblock Building end-to-end dialogue systems using generative hierarchical
  neural network models.
\newblock In {\em AAAI}.

\bibitem[\protect\citeauthoryear{Serban \bgroup et al\mbox.\egroup
  }{2017}]{DBLP:journals/corr/SerbanSLCPCB16}
Serban, I.~V.; Sordoni, A.; Lowe, R.; Charlin, L.; Pineau, J.; Courville,
  A.~C.; and Bengio, Y.
\newblock 2017.
\newblock A hierarchical latent variable encoder-decoder model for generating
  dialogues.
\newblock In {\em AAAI}.

\bibitem[\protect\citeauthoryear{Shang \bgroup et al\mbox.\egroup
  }{2018}]{DBLP:conf/ijcai/ShangFPFZY18}
Shang, M.; Fu, Z.; Peng, N.; Feng, Y.; Zhao, D.; and Yan, R.
\newblock 2018.
\newblock Learning to converse with noisy data: Generation with calibration.
\newblock In {\em IJCAI}.

\bibitem[\protect\citeauthoryear{Tay \bgroup et al\mbox.\egroup
  }{2019}]{DBLP:journals/acl/TayYi2019}
Tay, Y.; Wang, S.; Tuan, L.~A.; Fu, J.; Phan, M.~C.; Yuan, X.; Rao, J.; Hui,
  S.~C.; and Zhang, A.
\newblock 2019.
\newblock Simple and effective curriculum pointer-generator networks for
  reading comprehension over long narratives.
\newblock In {\em ACL}.

\bibitem[\protect\citeauthoryear{Tsvetkov \bgroup et al\mbox.\egroup
  }{2016}]{DBLP:conf/acl/TsvetkovFLMD16}
Tsvetkov, Y.; Faruqui, M.; Ling, W.; MacWhinney, B.; and Dyer, C.
\newblock 2016.
\newblock Learning the curriculum with bayesian optimization for task-specific
  word representation learning.
\newblock In {\em ACL}.

\bibitem[\protect\citeauthoryear{Vaswani \bgroup et al\mbox.\egroup
  }{2017}]{DBLP:conf/nips/VaswaniSPUJGKP17}
Vaswani, A.; Shazeer, N.; Parmar, N.; Uszkoreit, J.; Jones, L.; Gomez, A.~N.;
  Kaiser, L.; and Polosukhin, I.
\newblock 2017.
\newblock Attention is all you need.
\newblock In {\em NIPS}.

\bibitem[\protect\citeauthoryear{Weinshall, Cohen, and
  Amir}{2018}]{DBLP:conf/icml/WeinshallCA18}
Weinshall, D.; Cohen, G.; and Amir, D.
\newblock 2018.
\newblock Curriculum learning by transfer learning: Theory and experiments with
  deep networks.
\newblock In {\em ICML}.

\bibitem[\protect\citeauthoryear{Williams}{1992}]{DBLP:journals/ml/Williams92}
Williams, R.~J.
\newblock 1992.
\newblock Simple statistical gradient-following algorithms for connectionist
  reinforcement learning.
\newblock {\em Machine Learning} 8:229--256.

\bibitem[\protect\citeauthoryear{Xing \bgroup et al\mbox.\egroup
  }{2017}]{DBLP:conf/aaai/XingWWLHZM17}
Xing, C.; Wu, W.; Wu, Y.; Liu, J.; Huang, Y.; Zhou, M.; and Ma, W.
\newblock 2017.
\newblock Topic aware neural response generation.
\newblock In {\em AAAI}.

\bibitem[\protect\citeauthoryear{Xu \bgroup et al\mbox.\egroup
  }{2018}]{DBLP:conf/emnlp/XuDKR18}
Xu, X.; Dusek, O.; Konstas, I.; and Rieser, V.
\newblock 2018.
\newblock Better conversations by modeling, filtering, and optimizing for
  coherence and diversity.
\newblock In {\em EMNLP}.

\bibitem[\protect\citeauthoryear{Young \bgroup et al\mbox.\egroup
  }{2018}]{DBLP:conf/aaai/YoungCCZBH18}
Young, T.; Cambria, E.; Chaturvedi, I.; Zhou, H.; Biswas, S.; and Huang, M.
\newblock 2018.
\newblock Augmenting end-to-end dialogue systems with commonsense knowledge.
\newblock In {\em AAAI}.

\bibitem[\protect\citeauthoryear{Zhang \bgroup et al\mbox.\egroup
  }{2018a}]{DBLP:conf/acl/KielaWZDUS18}
Zhang, S.; Dinan, E.; Urbanek, J.; Szlam, A.; Kiela, D.; and Weston, J.
\newblock 2018a.
\newblock Personalizing dialogue agents: {I} have a dog, do you have pets too?
\newblock In {\em ACL}.

\bibitem[\protect\citeauthoryear{Zhang \bgroup et al\mbox.\egroup
  }{2018b}]{DBLP:conf/nips/ZhangGGGLBD18}
Zhang, Y.; Galley, M.; Gao, J.; Gan, Z.; Li, X.; Brockett, C.; and Dolan, B.
\newblock 2018b.
\newblock Generating informative and diverse conversational responses via
  adversarial information maximization.
\newblock In {\em NeurIPS}.

\bibitem[\protect\citeauthoryear{Zhang \bgroup et al\mbox.\egroup
  }{2019}]{DBLP:journals/acl19/hainanzhang}
Zhang, H.; Lan, Y.; Pang, L.; Guo, J.; and Cheng, X.
\newblock 2019.
\newblock Recosa: Detecting the relevant contexts with self-attention for
  multi-turn dialogue generation.
\newblock In {\em ACL}.

\bibitem[\protect\citeauthoryear{Zhao, Zhao, and
  Esk{\'{e}}nazi}{2017}]{DBLP:conf/acl/ZhaoZE17}
Zhao, T.; Zhao, R.; and Esk{\'{e}}nazi, M.
\newblock 2017.
\newblock Learning discourse-level diversity for neural dialog models using
  conditional variational autoencoders.
\newblock In {\em ACL}.

\bibitem[\protect\citeauthoryear{Zhou \bgroup et al\mbox.\egroup
  }{2018}]{DBLP:conf/aaai/ZhouHZZL18}
Zhou, H.; Huang, M.; Zhang, T.; Zhu, X.; and Liu, B.
\newblock 2018.
\newblock Emotional chatting machine: Emotional conversation generation with
  internal and external memory.
\newblock In {\em AAAI}.

\end{thebibliography}

\end{document}